\documentclass{article}
\usepackage{spconf,amsmath,graphicx}

\usepackage{amsmath,amssymb} % define this before the line numbering.
\usepackage{color}
\usepackage{balance}
\usepackage{amsmath}
\usepackage[linesnumbered,ruled,vlined]{algorithm2e}
\usepackage{algpseudocode}% http://ctan.org/pkg/algorithmicx
\usepackage{enumitem}% http://ctan.org/pkg/enumitem


\usepackage{graphicx}
\usepackage{pdfpages}
\usepackage{ragged2e}
\usepackage[subtle]{savetrees}

\graphicspath{{figures/}} 
\newcommand{\etal}{\textit{et al}.}

\title{Mask-based Data Augmentation for Semi-supervised Semantic Segmentation}
\name{Ying Chen, Xu Ouyang, Kaiyue Zhu, Gady Agam}
\address{Illinois Institute of Technology\\
				Department of Computer Science \\
				Chicago, IL 60616, USA\\
				ychen245, xouyang3, kzhu6@hawk.iit.edu, agam@iit.edu}
\begin{document}
%\ninept
%
\maketitle
\begin{abstract}
Semantic segmentation using convolutional neural networks (CNN) is a crucial component in image analysis. Training a CNN to perform semantic segmentation requires a large amount of labeled data, where the production of such labeled data is both costly and labor intensive. Semi-supervised learning algorithms address this issue by utilizing unlabeled data and so reduce the amount of labeled data needed for training.
In particular, data augmentation techniques such as CutMix and ClassMix generate additional training data from existing labeled data. In this paper we propose a new approach for data augmentation, termed ComplexMix, which incorporates aspects of CutMix and ClassMix with improved performance. The proposed approach has the ability to control the complexity of the augmented data while attempting to be semantically-correct and address the tradeoff between complexity and correctness.
The proposed ComplexMix approach is evaluated on a standard dataset for semantic segmentation and compared to other state-of-the-art techniques. Experimental results show that our method yields improvement over state-of-the-art methods on standard datasets for semantic image segmentation. 
% which is a good tradeoff between
\end{abstract}


\begin{figure*}[h]

\begin{minipage}[b]{1.0\linewidth}
  \centering
  \centerline{\includegraphics[scale=0.4]{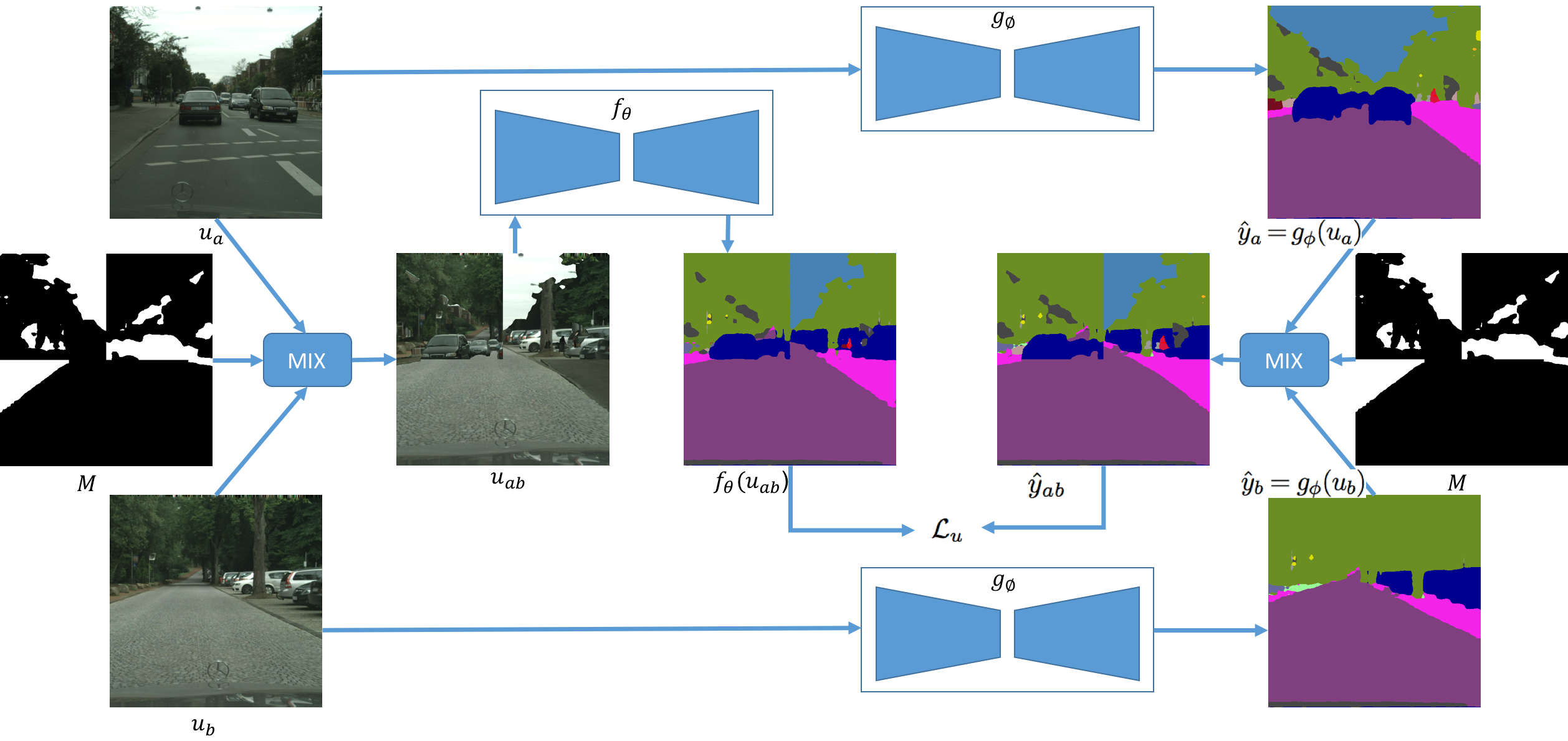}} %width=\textwidth,height=8cm
 %\vspace{2.0cm}
\end{minipage}
\caption{Illustration of our proposed approach to semi-supervised segmentation via mask-based data augmentation. Our approach uses the mean teacher strategy. The top and bottom branches in this network belong to the teacher who is trained to produce semantic segmentation predictions, whereas the middle branch belongs to the student which attempts to match mixed predictions from the teacher with its own predictions based on the mixed image input.}
\label{fig:arc}
\end{figure*}

\section{Introduction}

Semantic segmentation is concerned with assigning a semantic label to pixels belonging to certain objects in an image. Semantic segmentation is fundamental to image analysis and serves as a high-level pre-processing step to support many applications including scene understanding and autonomous driving. CNN-based fully-supervised approaches have achieved remarkable results in semantic segmentation of standard datasets. Generally, when sufficient labeled data is available, training a state-of-the-art network can easily achieve high accuracy. Labeling a large set of samples is expensive and time consuming and so the goal in semi-supervised semantic segmentation is to use a small labeled set and a large unlabeled set to train the network thus reducing the amount of labeled data needed.

Consistency regularization has been applied to semi-supervised classification~\cite{oliver2018realistic, miyato2018virtual, sohn2020fixmatch} yielding significant progress in the past few years. The key idea behind consistency regularization is to apply various data augmentations to encourage consistent predictions for unlabeled samples. Its effectiveness relies on the observation that the decision boundary of a classifier usually lies in low density regions and so can benefit from clusters formed by augmented data~\cite{chapelle2005semi}. While consistency regularization has been successfully employed for classification tasks, applying traditional data augmentation techniques to semantic segmentation has been shown~\cite{french2019semi} to be less effective as semantic segmentation may not exhibit low density regions around class boundaries. Several approaches have been developed to address this issue by applying augmentation on encoded space instead of input space~\cite{ouali2020semi}, or by enforcing consistent predictions for unsupervised mixed samples as in CutMix~\cite{yun2019cutmix, french2019semi}, CowMix~\cite{french2020milking}, and ClassMix~\cite{olsson2020classmix}. 

The method proposed in this paper belongs to the category of enforcing consistent predictions for unsupervised mixed samples. We propose a more effective mask-based augmentation strategy for segmentation maps, termed ComplexMix, to address semi-supervised semantic segmentation. We hypothesize that there is added value in increasing the complexity of semantically correct augmentation and so attempt to produce complex augmentation which is semantically correct. We do so by splitting the segmentation map of one image into several squares of identical size and predict semantic labels in each square based on the current model. Following the augmentation strategy of ClassMix~\cite{olsson2020classmix}, we then select in each square half of the predicted classes and paste them onto the augmented image to form a new augmentation that respects semantic boundaries. The complexity of the augmentation is controlled by the number of squares generated in the initial split.
Experimental evaluation results demonstrate that the proposed ComplexMix augmentation is superior to random augmentations or simple semantically correct augmentation techniques.

The key contribution of this paper is in employing consistency regularization to semantic segmentation through a novel data augmentation strategy for producing complex and semantically-correct data from unlabeled examples. The proposed approach has the ability to control the complexity of the augmented data and so balance a tradeoff between complexity ad correctness. Experimental evaluation results on a standard dataset 
demonstrate improved performance over state-of-the-art techniques.

%can be summarized as follows: 
%\begin{itemize}
% \item We propose a simple yet effective mask-based augmentation method, which is a good tradeoff between complexity of the boundary and semantics of the objects comparing with CutMix and ClassMix.
% \item Our method aims to avoid large objects dominating the mixing on some level.
  %\item The proposed method encourages mixed objects to have more complex local features. 
%  \item We demonstrate that our approach gets improvement over state of the art methods on standard datasets.
%\end{itemize}

\section{Related work}
Semi-supervised semantic segmentation has been studied using different mechanisms, including generative adversarial learning~\cite{hung2018adversarial,mittal2019semi}, pseudo labeling~\cite{feng2020semi,chen2020leveraging}, and consistency regularization~\cite{french2019semi,kim2020structured,olsson2020classmix}. 

%\subsection{Generative adversarial learning}
\medskip\noindent
\textbf{Generative adversarial learning}.
GAN-based adversarial learning has been applied to semi-supervised semantic segmentation in different ways. Mittal \etal~\cite{mittal2019semi} use two network branches to link semi-supervised classification with semi-supervised segmentation, including self-training, and so reduce both low- and the high-level artifacts typical when training with few labels. In~\cite{hung2018adversarial}, fully convolutional discriminator enables semi-supervised learning through discovering trustworthy regions in predicted results of unlabeled images, thereby providing additional supervisory signal.

%Mittal \etal~\cite{mittal2019semi} and Hung \etal~\cite{hung2018adversarial} employ a discriminator in a fully convolutional manner to differentiate the predicted probability maps from the ground truth segmentation distribution to guide learning. 

%\subsection{Pseudo labeling}
\medskip\noindent
\textbf{Pseudo labeling}.
Pseudo labeling is a commonly used technique for semi-supervised learning in semantic segmentation. Feng \etal~\cite{feng2020semi} exploit inter-model disagreement based on prediction confidence to construct a dynamic loss which is robust against pseudo label noise, and so enable it to extend pseudo labeling to class-balanced curriculum learning. Chen \etal~\cite{chen2020leveraging} predict pseudo-labels for unlabeled data and train subsequent models with both manually-annotated and pseudo-labeled data.

%This technique has exhibited its great success in semi-supervised semantic segmentation~\cite{feng2020semi,chen2020leveraging}. 
%Pseudo-Labeling is closely related to entropy minimization, encouraging high-confident predictions for unlabeled samples produced by the labeled data model. 

%\subsection{Consistency regularization}
\medskip\noindent
\textbf{Consistency regularization}.
Consistency regularization works by enforcing a learned model to produce robust predictions for perturbations of unlabeled samples. Consistency regularization for semantic segmentation was first successfully used for medical imaging but has since been applied to other domains. French \etal~\cite{french2019semi} attribute the challenge in semi-supervised semantic segmentation to cluster assumption violations, and propose a data augmentation technique termed CutMix~\cite{yun2019cutmix} to solve it. Ouali \etal~\cite{ouali2020semi} apply perturbations to the output of an encoder to preserve the cluster assumption. Olsson \etal~\cite{olsson2020classmix} propose a similar technique based on predictions by a segmentation network to construct mixing, thus encouraging consistency over highly varied mixed samples while respecting semantic boundaries in the original images. Our proposed method incorporates ideas from~\cite{french2019semi} and~\cite{olsson2020classmix} to enforce a tradeoff between complexity and correctness and avoid the problem where large objects dominate the mixing.
%While French \etal~\cite{french2019semi} propose to employ a data augmentation technique called CutMix~\cite{yun2019cutmix}, a mask-based mixing method where randomized rectangular regions are cut out from one image and pasted onto another, to avoid this cluster. 

%-------------------------------------------------------------------------
\section{Proposed semi-supervised learning approach}
In this section, we present our proposed approach for addressing semi-supervised semantic segmentation. We introduce the proposed augmentation strategy termed ComplexMix, discuss the loss functions used to guide the model parameter estimate, and provide details of the training procedure.

% \begin{table}
%\caption{Evaluation results($\%$). The symbol $-$ indicates data not provided in reference paper. (*) means the baseline results reported in [10,11,5,12] have small variations which are omitted here for brevity}
%\begin{center}
%\begin{tabular}{llllll} 
%  \hline
%  Labeled samples & 1/30 & 1/8 & 1/4  & 1/2 & Full(2975)\\
%  \hline
%  Baseline         &-  &55.5 &59.9 &64.1 &66.4\\
%  Adversarial~\cite{hung2018adversarial}   &-  &58.8 &62.3 &65.7 &-\\
%  Improvement        &- &3.3 &2.4 &1.6 &-\\
%  \hline
%  
%  Baseline         &-  &56.2 &60.2 &- &66.0\\
%  s4GAN~\cite{mittal2019semi}         &-  &59.3 &61.9 &- &65.8\\
%  Improvement        &-  &3.1 &1.7 &- &-0.2\\
%  \hline
%  
%  Baseline         &44.41  &55.25 &60.57 &- &67.53\\
%  French \etal~\cite{french2019semi}         &51.20  &60.34 &63.87 &- &-\\
%  Improvement        &6.79  &5.09 &3.3 &- &-\\
%  \hline
%  
%   Baseline         &45.5  &56.7 &61.1 &- &66.9\\
%  DST-CBC~\cite{feng2020semi}         &48.7  &60.5 &\textbf{64.4}  &- &-\\
%  Improvement        &3.2 &3.8 &3.3 &- &-\\
%  \hline
%  
%   Baseline         &43.84  &54.84 &60.08 &63.02 &66.19\\
%  ClassMix~\cite{olsson2020classmix}         &\textbf{54.07}  &61.35 &63.63 &66.29 &-\\
%  Improvement        &10.23  &6.51 &3.72 &3.27 &-\\
%  \hline
%  
% Baseline(*)         &43.84  &54.84 &60.08 &63.02 &66.19\\
%  Ours         &53.88  &\textbf{62.25} &64.07 &\textbf{66.77} &N/A\\
%  Improvement        &10.04  &\textbf{7.41} &3.99 &\textbf{3.75} &N/A\\
%  \hline
%\end{tabular}
%\label{table:evalres}
%\end{center}
%\end{table}

\subsection{ComplexMix for semantic segmentation}

%\subsubsection{Mean-teacher framework}
\medskip\noindent
\textbf{Mean-teacher framework}.
The proposed approach follows commonly employed state-of-the-art semi-supervised learning techniques~\cite{french2019semi,french2020milking,olsson2020classmix} by using the mean teacher framework~\cite{tarvainen2017mean}, where the student and teacher networks have identical structure. In this approach the student network is updated by training whereas the teacher network is updated by blending its parameters with that of the student network. 
Our approach follows interpolation consistency training (ICT)~\cite{verma2019interpolation} by feeding an input image pair to the teacher network and a blended image to the student network. We then enforce correspondence between student predictions on blended input and blended teacher predictions. An illustration of this framework is shown in Figure~\ref{fig:arc}. 
In this figure, the student and teacher segmentation networks are denoted by $f_{\theta}$ and $g_{\phi}$, respectively, where $\theta$ and $\phi$ are the network parameters. The input image pair to be mixed is denoted by $u_a$ and $u_b$, and the mixed image is denoted by $u_{ab}$. The blending mask used to generate the mix is denoted by $M$. To generate the mask $M$ the teacher provides predictions $\hat{y}_a=g_{\theta}(u_a)$ and $\hat{y}_b=g_{\theta}(u_b)$. The teacher's mixed prediction for $u_{ab}$ is denoted by $\hat{y}_{ab}$ whereas the student's prediction for $u_{ab}$ is given by $f_{\theta}(u_{ab})$. The consistency loss term enforcing correspondence between student and blended teacher predictions is denoted by $\mathcal{L}_{u}$. All the data used in this figure is unsupervised.
%========= check figure: L_semi, \hat_y
%
%Preliminary experiments demonstrate that applying mean-teacher framework yields good performance in semantic segmentation. Thus, our work also employs this framework in our experiments, using $f_{\theta}$ and $g_{\phi}$ to denote student (segmentation) network with parameters $\theta$ and teach network with parameters $\phi$ respectively.

%\subsubsection{Mixing strategy}
\medskip\noindent
\textbf{Mixing strategy}.
\label{sec:mixing-strategy}
Producing a mix of images for training the student is possible in different ways. The proposed approach uses a mask $M$ to achieve this. Given a pair of images ($u_a$, $u_b$) and a mask $M$, a portion of $u_a$ defined by $M$ can be cut from $u_a$ and pasted onto $u_b$ to create a mixed image $u_{ab}=M\odot u_a +(1-M)\odot u_b$.
% where $\odot$ represents Hadamard product. 
Likewise, semantic labels $\hat{y}_a$ and $\hat{y}_b$ could be mixed using $M$ to produce the mixed semantic label $\hat{y}_{ab}$.
Different approaches for generating the mask $M$ exist. The proposed ComplexMix strategy combines ideas from CutMix and ClassMix to generate $M$.
%\subsubsection{CutMix}
In CutMix~\cite{yun2019cutmix, french2019semi}
%(combining aspects of MixUp and CutOut) 
the mask $M$ is a random rectangular region with area covering half of the image.
%use mean square error as the consistency loss. 
%
%\subsubsection{ClassMix}
In ClassMix~\cite{olsson2020classmix}, the mask $M$ is generated based on semantic labels produced by a network. 
%Given semantic labels for $C$ classes, $C/2$ classes are randomly selected and the binary mask $M$ is created using the pixels belonging to these classes.
%
%Two unlabeled images $x_{a}$ and $x_{b}$ are randomly sampled from the unlabeled dataset. Then $x_{a}$ and $x_{b}$ are fed into the segmentation network trained on labeled set, producing the predictions $g_{\phi}(x_{a})$ and $g_{\phi}(x_{b})$. Next, the half number of the classes are randomly selected from the argmaxed prediction of $x_{a}$ to generate the mask $M$, filled with value $1$ in chosen pixels while filled with value $0$ otherwise. After that, the mixed image $x_{ab}$ is produced by involving pixels from $x_{a}$ for value $1$ in $M$ and pixels from $x_{b}$ elsewhere. The pseudo label for $x_{ab}$ based on the predictions of $x_{a}$ and $x_{b}$ is produced in the same way. 

%\subsubsection{ComplexMix}
The motivation for the proposed ComplexMix strategy is to create complex and semantically-correct mixing masks $M$.
% by incorporating aspects from CutMix and ClassMix. 
%
Given two images $u_a$ and $u_b$ with corresponding semantic labels $\hat{y}_a$ and $\hat{y}_b$, we split $u_a$ and its corresponding semantic label $\hat{y}_a$ into $p\times p$ equal size blocks. In each block we randomly select $C/2$ classes (where $C$ is the total number of classes) and use the pixels belonging to the selected classes (based on $\hat{y}_a$) to form the mask $M$.

The parameter $p$ is used to control the complexity of the mask. With a higher value of $p$ there are more blocks and so we have a more granular mixing with higher complexity. However, because the boundaries of blocks are arbitrary they introduce errors into the mixing. There is, thus, a tradeoff between complexity and correctness that needs to be balanced by the selection of the parameter $p$. In our experiments we treat $p$ as a hyper parameter and determine its value empirically.
The selection of the parameter $p$ may depend on the size of objects in the image (a larger $p$ is possible for small objects). Subsequently, to account for different scales of objects in the image, instead of a fixed value for $p$ we select it randomly during each iteration from a possible set of values ($[4, 16, 64, 128]$ in our experiments).

There are three key benefits to the proposed ComplexMix strategy: preventing large objects from dominating the blended image, forcing mixed objects to have more complex boundaries, and  controlling the tradeoff between complexity and correctness.

%\subsubsection{Algorithm}
\medskip\noindent
\textbf{Algorithm}.
The student model $f_{\theta}$ is initially trained based on labeled data using a supervised segmentation loss. The teacher model is then initialized by copying the student network weights. Note that the student and teacher networks are identical.
We denote the supervised training set using $S=\{(s,y)|s\in {R^{H \times W \times 3}}, y \in {(1,C)^{H \times W}}\}$, where each sample $s$ is an $H \times W$ color image which is associated with a ground-truth C-class segmentation map $y$. Each entry $y^{i,j}$ takes a class label from a finite set (${1,2,...,C}$) or a one-hot vector $[y^{(i,j,c)}]_{c}$.
Similarly, we denote the unlabeled set using $U=\{(u)|u\in {R^{H \times W \times 3}}\}$. 

After the initial training of the student using supervised data, the training continues using both supervised and unsupervised data. Two images $u_{a}$ and $u_{b}$ are randomly sampled from the unlabeled dataset $U$ and fed into the teacher's model $g_{\phi}$ to produce pseudo-labels (segmentation map predictions) $\hat{y}_a=g_{\phi}(u_a)$ and $\hat{y}_b=g_{\phi}(u_b)$. To improve performance we use a common self-supervised-learning (SSL) method where pseudo-labels are assigned to to unlabeled samples only when the confidence in the label is sufficiently high.
The pseudo-labels are then used to produce a mixing mask $M$ and a mixed image $u_{ab}$ with a corresponding pseudo-label $\hat{y}_{ab}$ as described in Section \ref{sec:mixing-strategy}. The pseudo label $\hat{y}_{ab}$ is used to train the student through the unsupervised loss term $\mathcal{L}_{u}$.
In addition, supervised images $s$ are selected from the labeled set $S$ and used to train the student through the supervised loss term $\mathcal{L}_{s}$.

%Specifically, based on the prediction probability produced for the unlabeled samples, we could use SSL to obtain pseudo labels $\hat{y}$ with high confidence, using a fixed or scheduled threshold. In our work, we use the same value as ClassMix.

 \begin{table*}
\caption{Evaluation results showing mean IoU in percent for different portions of the data with labels. The symbol ``-'' indicates data was not provided in reference paper. The different columns show the fraction of labeled data used in training.
}
\begin{center}
\begin{tabular}{lllllll} 
  \hline
  Group & Labeled samples & 1/30 & 1/8 & 1/4  & 1/2 & Full\\
  \hline
  \hline
  1 & Deeplab-V2         &43.84  &54.84 &60.08 &63.02 &66.19\\
  \hline
  2 & Adversarial~\cite{hung2018adversarial}   &-  &58.8 &62.3 &65.7 &N/A\\
  %\hline
  & s4GAN~\cite{mittal2019semi}         &-  &59.3 &61.9 &- &N/A\\
  %\hline
  & DST-CBC~\cite{feng2020semi}         &48.7  &60.5 &\textbf{64.4}  &- &N/A\\
  \hline
  3 & French \etal~\cite{french2019semi}         &51.20  &60.34 &63.87 &- &N/A\\
  %\hline
  & ClassMix~\cite{olsson2020classmix}         &\textbf{54.07}  &61.35 &63.63 &66.29 &N/A\\
  \hline
  4 & \textbf{Ours (ComplexMix)}        &53.88 $\pm$  0.56 &\textbf{62.25} $\pm$ 1.22 &64.07 $\pm$ 0.46 &\textbf{66.77} $\pm$ 0.83 &N/A\\
  \hline
\end{tabular}
\label{table:evalres}
\end{center}
\end{table*}

\subsection{Loss function and training}

%\subsubsection{Loss function}
\medskip\noindent
\textbf{Loss function}.
Our model is trained to minimize a combined loss composed of a supervised loss term $\mathcal{L}_{s}$ and an unsupervised consistency loss term $\mathcal{L}_{u}$: 
%%%designed to enforce consistency between the student and teacher classifiers:
\begin{equation}
\mathcal{L}=\mathcal{L}_{s}(f_{\theta}(s),y)+\lambda\mathcal{L}_{u}(f_{\theta}(u),g_{\phi}(u))
\end{equation}
In this equation $\lambda$ is a hyper-parameter used to control the balance between the supervised and unsupervised terms.

The supervised loss term $\mathcal{L}_{s}$ is used to train the 
student model $f_{\theta}$ with labeled images in a supervised manner using the categorical cross entropy loss:
\begin{equation}
\mathcal{L}_{s}(f_{\theta}(s),y)=-\dfrac{1}{N}\sum_{i=1}^{N}\sum_{j=1}^{H \times W}\sum_{c=1}^{C} y^{(i,j,c)}\log f_{\theta}(s)^{(i,j,c)}
\label{eq:seg1}
\end{equation}
where $N$ is the total number of labeled examples. In this equation, $y^{(i,j,c)}$ and $f_{\theta}(s)^{(i,j,c)}$ are the target and predicted probabilities for pixel $(i, j)$ belonging to class $c$, respectively.

The unsupervised loss term $\mathcal{L}_{u}$ is used to train the
student model $f_{\theta}$ with unlabeled image pairs $u_{a}$ and $u_{b}$ using the categorical cross entropy loss to match pseudo labels:
\begin{equation}
\mathcal{L}_{u}(f_{\theta}(u_{ab}),\hat{y}_{ab})=-\dfrac{1}{N}\sum_{i=1}^{N}\sum_{j=1}^{H \times W}\sum_{c=1}^{C} \hat{y}_{ab}^{(i,j,c)}\log f_{\theta}(u_{ab})^{(i,j,c)}
\label{eq:seg2}
\end{equation}
where $u_{ab}$ is the mixed image of $u_{a}$ and $u_{b}$ using $M$, and $\hat{y}_{ab}$ is the mixed pseudo label of $\hat{y}_a=g_{\phi}(u_{a})$ and $\hat{y}_b=g_{\phi}(u_{b})$ based on $M$.
As described earlier, the teacher model is updated by blending its coefficients with updated student coefficients.

%\subsubsection{Training details}
\medskip\noindent
\textbf{Training details}.
To obtain high-quality segmentation results, it is critical to choose a strong base model. In this work, we use Deeplab-V2~\cite{chen2017deeplab} with a pretrained ResNet-101~\cite{he2016deep} model, as the base semantic segmentation network $f_{\theta}$. 

%pretrained on ImageNet~\cite{deng2009imagenet} and MSCOCO~\cite{lin2014microsoft}, 

We use the Pytorch deep learning framework to implement our network on two NVIDIA-SMI GPU with $16$ GB memory in total. Stochastic Gradient Descent is employed as the optimizer with momentum of $0.9$ and a weight decay of $5 \times 10^{-4}$ to train the model. The initial learning rate is set to $2.5 \times 10^{-4}$ and decayed using the polynomial decay schedule of~\cite{chen2017deeplab}.

%-------------------------------------------------------------------------

\section{Evaluation}
In this section, we present experimental results using common metrics. We evaluate the proposed approach and compare it with known approaches using standard evaluation datasets.

%\subsection{Datasets}
\medskip\noindent
\textbf{Datasets}.
We demonstrate the effectiveness of our method on the standard Cityscapes~\cite{cordts2016cityscapes} urban scene dataset. The dataset consists of $2975$ training images and $500$ validation images. We follow the common standard of the baseline methods we compare to by resizing each image to $512 \times 1024$ without any additional augmentation such as random cropping or scaling. The batch size for labeled and unlabeled samples is set to $2$ for training, and the total number of training iterations is set to $40k$ following the settings in~\cite{hung2018adversarial, olsson2020classmix}.

%\subsection{Evaluation metrics}
\medskip\noindent
\textbf{Evaluation metrics}.
To evaluate the proposed method, we use Intersection over Union (IoU) which is a commonly used metric for semantic segmentation. The different columns show the fraction of labeled data used in training.
%The metric is calculated with true positive (TP), false positive (FP), true negative (TN), and false negative (FN). 
When training on fraction of the data we repeated each experiment $5$ times and computed the average IoU value of all experiments for all classes in the dataset.

%\subsection{Results}
\medskip\noindent
\textbf{Results}.
The evaluation results for Cityscapes are shown in Table~\ref{table:evalres} where entries indicate mean  intersection-over-union (mIoU) percentages. A higher mIoU indicates better results. The different columns show the fraction of labeled data used in training. We compare the proposed approach with six baseline methods, all using the same DeepLab-v2 framework. 
The baseline result in group 1 is a fully supervised method that does not take advantage of unlabeled data. It is a lower bound for results. 
The methods in group 2 are semi-supervised approaches using unlabeled data in an adversarial way. 
The methods in group 3 use mask-based data augmentation, and are in the same category as the proposed approach.
N/A indicates the full labeled data set is used for supervised learning, while ``-'' indicates the evaluation was not reported in the reference paper.
Note that the baselines Deeplab-V2 results reported  in~\cite{hung2018adversarial,mittal2019semi,french2019semi,olsson2020classmix} have small insignificant variations compared with the results shown here.

As can be expected, smaller portions of labeled data result in reduced performance. However, observe in the table that adding unlabeled data with semi-supervised approaches improves performance in a meaningful way. The methods in group 3 where augmentation is used generally preform better than the methods in group 2. The proposed ComplexMix approach belongs to the the class of group 3 and as can be observed obtains results which are better than other group 3 methods in most cases.

%Since our work is built upon~\cite{olsson2020classmix}, we use its baseline results as the baseline results in the table. 
%As observed in Table~\ref{table:evalres}, the proposed method, incorporating ideas from CutMix~\cite{french2019semi} and ClassMix~\cite{olsson2020classmix} , realizes improvements over both of them in almost all cases. Thus, we believe our method has a better tradeoff between complexity of the boundary and semantics of the objects than either CutMix or ClassMix, and also benefits from avoiding dominant large object on some level and encouraging mixing objects have more complex local features as well. Besides, our method also outperforms the two adversarial methods, Adversarial~\cite{hung2018adversarial} and s4GAN~\cite{mittal2019semi}, demonstrating the effectiveness of our method. 

%-------------------------------------------------------------------------
\section{Conclusion}
In this paper, we address the problem of semi-supervised learning for semantic segmentation using mask-based data augmentation. We propose a new augmentation technique that can balance between complexity and correctness and show that by using it we are able to improve on the state-of-the-art when evaluating semantic segmentation over a standard dataset.

%In future work, we plan to explore a new method to tackle this problem. 

% -------------------------------------------------------------------------
%\afterpage{\clearpage}
\bibliographystyle{IEEEbib}
\bibliography{ref}

\begin{thebibliography}{10}

\bibitem{oliver2018realistic}
Avital Oliver, Augustus Odena, Colin~A Raffel, Ekin~Dogus Cubuk, and Ian
  Goodfellow,
\newblock ``Realistic evaluation of deep semi-supervised learning algorithms,''
\newblock in {\em Advances in neural information processing systems}, 2018, pp.
  3235--3246.

\bibitem{miyato2018virtual}
Takeru Miyato, Shin-ichi Maeda, Masanori Koyama, and Shin Ishii,
\newblock ``Virtual adversarial training: a regularization method for
  supervised and semi-supervised learning,''
\newblock {\em IEEE transactions on pattern analysis and machine intelligence},
  vol. 41, no. 8, pp. 1979--1993, 2018.

\bibitem{sohn2020fixmatch}
Kihyuk Sohn, David Berthelot, Chun-Liang Li, Zizhao Zhang, Nicholas Carlini,
  Ekin~D Cubuk, Alex Kurakin, Han Zhang, and Colin Raffel,
\newblock ``Fixmatch: Simplifying semi-supervised learning with consistency and
  confidence,''
\newblock {\em arXiv preprint arXiv:2001.07685}, 2020.

\bibitem{chapelle2005semi}
Olivier Chapelle and Alexander Zien,
\newblock ``Semi-supervised classification by low density separation.,''
\newblock in {\em AISTATS}. Citeseer, 2005, vol. 2005, pp. 57--64.

\bibitem{french2019semi}
Geoff French, Timo Aila, Samuli Laine, Michal Mackiewicz, and Graham Finlayson,
\newblock ``Semi-supervised semantic segmentation needs strong,
  high-dimensional perturbations,''
\newblock {\em CoRR, abs/1906.01916}, 2019.

\bibitem{ouali2020semi}
Yassine Ouali, C{\'e}line Hudelot, and Myriam Tami,
\newblock ``Semi-supervised semantic segmentation with cross-consistency
  training,''
\newblock in {\em Proceedings of the IEEE/CVF Conference on Computer Vision and
  Pattern Recognition}, 2020, pp. 12674--12684.

\bibitem{yun2019cutmix}
Sangdoo Yun, Dongyoon Han, Seong~Joon Oh, Sanghyuk Chun, Junsuk Choe, and
  Youngjoon Yoo,
\newblock ``Cutmix: Regularization strategy to train strong classifiers with
  localizable features,''
\newblock in {\em Proceedings of the IEEE International Conference on Computer
  Vision}, 2019, pp. 6023--6032.

\bibitem{french2020milking}
Geoff French, Avital Oliver, and Tim Salimans,
\newblock ``Milking cowmask for semi-supervised image classification,''
\newblock {\em arXiv preprint arXiv:2003.12022}, 2020.

\bibitem{olsson2020classmix}
Viktor Olsson, Wilhelm Tranheden, Juliano Pinto, and Lennart Svensson,
\newblock ``Classmix: Segmentation-based data augmentation for semi-supervised
  learning,''
\newblock {\em arXiv preprint arXiv:2007.07936}, 2020.

\bibitem{hung2018adversarial}
Wei-Chih Hung, Yi-Hsuan Tsai, Yan-Ting Liou, Yen-Yu Lin, and Ming-Hsuan Yang,
\newblock ``Adversarial learning for semi-supervised semantic segmentation,''
\newblock {\em arXiv preprint arXiv:1802.07934}, 2018.

\bibitem{mittal2019semi}
Sudhanshu Mittal, Maxim Tatarchenko, and Thomas Brox,
\newblock ``Semi-supervised semantic segmentation with high-and low-level
  consistency,''
\newblock {\em IEEE Transactions on Pattern Analysis and Machine Intelligence},
  2019.

\bibitem{feng2020semi}
Zhengyang Feng, Qianyu Zhou, Guangliang Cheng, Xin Tan, Jianping Shi, and
  Lizhuang Ma,
\newblock ``Semi-supervised semantic segmentation via dynamic self-training and
  class-balanced curriculum,''
\newblock {\em arXiv preprint arXiv:2004.08514}, 2020.

\bibitem{chen2020leveraging}
Liang-Chieh Chen, Raphael~Gontijo Lopes, Bowen Cheng, Maxwell~D Collins, Ekin~D
  Cubuk, Barret Zoph, Hartwig Adam, and Jonathon Shlens,
\newblock ``Leveraging semi-supervised learning in video sequences for urban
  scene segmentation.,''
\newblock {\em arXiv preprint arXiv:2005.10266}, 2020.

\bibitem{kim2020structured}
Jongmok Kim, Jooyoung Jang, and Hyunwoo Park,
\newblock ``Structured consistency loss for semi-supervised semantic
  segmentation,''
\newblock {\em arXiv preprint arXiv:2001.04647}, 2020.

\bibitem{tarvainen2017mean}
Antti Tarvainen and Harri Valpola,
\newblock ``Mean teachers are better role models: Weight-averaged consistency
  targets improve semi-supervised deep learning results,''
\newblock {\em Advances in neural information processing systems}, vol. 30, pp.
  1195--1204, 2017.

\bibitem{verma2019interpolation}
Vikas Verma, Alex Lamb, Juho Kannala, Yoshua Bengio, and David Lopez-Paz,
\newblock ``Interpolation consistency training for semi-supervised learning,''
\newblock {\em arXiv preprint arXiv:1903.03825}, 2019.

\bibitem{chen2017deeplab}
Liang-Chieh Chen, George Papandreou, Iasonas Kokkinos, Kevin Murphy, and Alan~L
  Yuille,
\newblock ``Deeplab: Semantic image segmentation with deep convolutional nets,
  atrous convolution, and fully connected crfs,''
\newblock {\em IEEE transactions on pattern analysis and machine intelligence},
  vol. 40, no. 4, pp. 834--848, 2017.

\bibitem{he2016deep}
Kaiming He, Xiangyu Zhang, Shaoqing Ren, and Jian Sun,
\newblock ``Deep residual learning for image recognition,''
\newblock in {\em Proceedings of the IEEE conference on computer vision and
  pattern recognition}, 2016, pp. 770--778.

\bibitem{cordts2016cityscapes}
Marius Cordts, Mohamed Omran, Sebastian Ramos, Timo Rehfeld, Markus Enzweiler,
  Rodrigo Benenson, Uwe Franke, Stefan Roth, and Bernt Schiele,
\newblock ``The cityscapes dataset for semantic urban scene understanding,''
\newblock in {\em Proceedings of the IEEE conference on computer vision and
  pattern recognition}, 2016, pp. 3213--3223.

\end{thebibliography}

\end{document}